\pdfoutput=1

\documentclass[11pt]{article}

\usepackage[final]{acl}
\usepackage{booktabs}
\usepackage{times}
\usepackage{mathptmx} 
\usepackage{latexsym}
\usepackage{adjustbox}
\usepackage{amsmath}  
\usepackage{amssymb}  
\usepackage{multirow}
\usepackage{array}
\usepackage{graphicx} 
\usepackage{float} 
\usepackage{algorithm}
\usepackage{listings}
\usepackage{algpseudocode}
\usepackage{xspace}
\usepackage{graphicx}
\usepackage{etoolbox}
\usepackage{xcolor}
\usepackage{tcolorbox}
\usepackage{hyperref}
\usepackage{latexsym}

\usepackage{hyperref}

\usepackage[utf8]{inputenc} 
\usepackage[T1]{fontenc}    
\usepackage{hyperref}       
\usepackage{url}            
\usepackage{booktabs}       
\usepackage{amsfonts}       
\usepackage{nicefrac}       
\usepackage{microtype}      
\usepackage{xcolor}         
\usepackage{caption}
\usepackage{ragged2e}       
\usepackage{tcolorbox}
\usepackage{authblk}

\usepackage[T1]{fontenc}

\usepackage[utf8]{inputenc}
\usepackage{caption}

\tcbuselibrary{breakable}
\newtcolorbox{mytextbox}{
  breakable,
  colback=gray!10,
  colframe=black,
  boxrule=0.5pt,
  arc=4pt,
  left=6pt, right=6pt, top=6pt, bottom=6pt
}
\usepackage{microtype}

\usepackage{inconsolata}

\usepackage{graphicx}

\usepackage{microtype}
\newcommand{\system}{\emph{RAGPPI}\xspace}

\title{\system: Retrieval-Augmented Generation Benchmark for Protein–Protein Interactions in Drug Discovery}

\author{
  \textbf{Youngseung Jeon}$^{1}$ \quad
  \textbf{Ziwen Li}$^1$ \quad
  \textbf{Thomas Li}$^2$ \quad
  \textbf{JiaSyuan Chang}$^1$ \quad \\
  \textbf{Morteza Ziyadi}$^3$ \quad
  \textbf{Xiang `Anthony' Chen}$^{1}$\footnote{*} \\
  \vspace{0.2cm} \\
  $^1$University of California, Los Angeles \quad
  $^2$Palo Alto High School \quad
  $^3$Amazon AGI \\
  \vspace{0.1cm} \\
  \texttt{\{ysj, zil105, serenacjs, xac\}@g.ucla.edu} \\
  \texttt{thomasli13579@gmail.com, ziyadim@amazon.com} \\
  \vspace{0.1cm} \\
}

\newcommand\blfootnote[1]{%
  \begingroup
  \renewcommand\thefootnote{}\footnote{#1}%
  \addtocounter{footnote}{-1}%
  \endgroup
}

\begin{document}
\maketitle

\begin{abstract}
Retrieving the biological impacts of protein-protein interactions (PPIs) is essential for target identification (Target ID) in drug development. Given the vast number of proteins involved, this process remains time-consuming and challenging. Large Language Models (LLMs) and Retrieval-Augmented Generation (RAG) frameworks have supported Target ID; however, no benchmark currently exists for identifying the biological impacts of PPIs. To bridge this gap, we introduce the RAG Benchmark for PPIs (\system)~\footnote{\textcolor{blue}{Dataset link:} \url{https://huggingface.co/datasets/Youngseung/RAGPPI}}, a factual question-answer benchmark of 4,420 question-answer pairs that focus on the potential biological impacts of PPIs. Through interviews with experts, we identified criteria for a benchmark dataset, such as a type of QA and source. We built a gold-standard dataset (500 QA pairs) through expert-driven data annotation. 
We developed an ensemble auto-evaluation LLM~\footnote{\textcolor{blue}{Code link:} \url{https://github.com/youngseungjeon/RAGPPI}}\blfootnote{*Corresponding author.} that incorporates expert labeling characteristics, average fact–abstract similarity ($F_1$), and low-similarity fact counts ($F_2$), enabling the construction of a silver-standard dataset (3,720 QA pairs).
We are committed to maintaining \system as a resource to support the research community in advancing RAG systems for drug discovery QA solutions.

\blfootnote{Published in Proceedings of EACL 2026, licensed under CC BY 4.0; please cite the EACL version.}
\end{abstract}

\section{Introduction}

Drug discovery has the potential to yield new therapies that save lives. A historical example is Penicillin, which revolutionized the treatment of bacterial infections and prevented countless deaths since its development in the early 20th century~\cite{drews2000drug}. Despite its impact, the discovery process for new drugs remains lengthy and costly: typically spanning about a decade and costing approximately \$2 billion to bring a new drug to market~\cite{wouters2020estimated, hinkson2020accelerating}. One of the key early steps in this time- and cost-intensive process is identifying target proteins to bind to drugs, a process called \textit{ target identification} (Target ID). 

Target ID aims to identify protein-protein interactions (PPIs), which are protein pathways from an initial protein (IP) to protein candidates for the target protein (TP)~\cite{rasul2022target}. TP should have a biological, functional, or physical effect on IP when a drug binds to TP. These impacts ultimately contribute to therapeutic effects for a given disease. However, given that the number of protein candidates in the human body is several billion~\cite{smith2013proteoform}, \textit{Target ID} is very time-consuming and expensive, requiring researchers to explore PPI candidates within the extensive protein space by scanning related literature.


Recently, Large Language Models (LLMs)~\cite{openai2023chatgpt, touvron2023llama, devlin2019bert, google2024gemini} have emerged as powerful generative models and have been applied to this protein-related task, where LLMs infer potential biological impacts of protein-protein interactions (PPIs).
For example, 
Tx-LLM \cite{chaves2024tx} is a generalist large language model fine-tuned to encode knowledge across diverse therapeutic modalities and stages of the drug discovery pipeline.
JSL-MedLlama-8B \cite{johnsnowlabs2024medllama} is a fine-tuned model based on Llama-8B to support inference of medical QA tasks with a focus on medical genetics. 
However, they are prone to hallucinations that introduce unreliable results \cite{ji2023survey, zhang2023siren}.

\textit{Retrieval-Augmented Generation (RAG)} has emerged as a promising solution to alleviate this issue by combining a constantly updated database with efficient information retrieval for more accurate and contextually relevant responses \cite{lewis2020retrieval, li2023understand, khandelwal2019generalization}. The performance of RAG in improving the reliability of generative models \cite{martineau2023retrieval, zhang2023knowledge} has extended its application to Target ID, where reliability and explainability matter for domain experts~\cite{li2025grappi, xiong2024benchmarking, garigliotti2023retrieval}. 
Despite these advancements, the NLP community lacks a specialized benchmark to evaluate RAG systems on scientific factuality and reasoning depth. This gap limits the potential applicability of RAG models for scientific discovery. 



We strive to build a benchmark where researchers evaluate their RAG in generating potential biological, functional, or physical effects of PPIs. We present a qualified benchmark dataset for RAG for Target ID through interviews and expert-driven data annotation with 18 domain experts. To further enhance its practical relevance, we developed an automatic evaluation model that captures expert labeling characteristics, thereby demonstrating the benchmark's applicability in real-world biomedical and AI research. Specifically, \system makes three contributions. 


\begin{itemize}

\item The first contribution is to derive the criteria for creating the dataset through an interview with experts. 
Through this user study, we identified the essential information required in the protein-protein interaction (PPI) analysis process and established a unified query template. This template is strategically designed to demand a multi-step reasoning chain—from (1) Entity Identification to (2) Mechanism Elucidation and (3) Therapeutic Impact.
Using BioGrid~\footnote{https://thebiogrid.org/}, the most widely adopted dataset for protein-protein interactions (PPIs), we constructed a benchmark dataset to evaluate the potential biological impacts of PPIs, strategically balancing literature frequency and interaction types. 
This expert-grounded formulation not only reflects real-world Target ID workflows but also challenges NLP models to perform precise information synthesis.
Ultimately, our systematic process of identifying domain-specific requirements provides a robust methodological framework for building benchmarks for scientific discovery in NLP.

\item The second contribution is the benchmark dataset itself. We constructed \system, a factual QA benchmark consisting of 4,420 question-answer pairs on the biological impacts of PPIs. Specifically, through expert-driven data annotation, we first created a gold-standard dataset of 500 PPIs, each annotated with multiple biological steps of downstream processes that ultimately contribute to therapeutic effects for a given disease. Furthermore, leveraging an automatic evaluation model, we generated a silver-standard dataset of 3,720 PPIs, resulting in a total of 4,420 annotated examples. To the best of our knowledge, \system~is the first benchmark specifically designed to evaluate Retrieval-Augmented Generation (RAG) models for Target ID.

\item The third contribution is a method to scale the dataset synthetically.
We developed an ensemble auto-evaluation LLM consisting of three LLMs, each specialized to capture features of expert labeling on ``Correct'' and ``Incorrect'' answers. By using an atomic fact-based method~\cite{min2023factscore}, we identified two key features associated with factual inconsistency:
1) average cosine similarity of atomic facts of answers to references, and 2) the number of lower outliers of atomic facts in cosine similarity to references.
The ensemble model determines the final label by aggregating predictions from sub-models through majority voting, achieving an accuracy of 93.71\%. This enables researchers to leverage our benchmarking dataset without requiring further expert involvement, thereby facilitating broader applicability across the biomedical and NLP communities.

\end{itemize}

\section{Task Formulation Grounded in Expert Interviews}
A RAG \textit{QA} system generates an answer \textit{A} to a given question \textit{Q} by leveraging both retrieved external sources and the model's internal knowledge. 
Domain-specific RAG systems require careful consideration of both QA types and retrieval references. To evaluate such a system in the context of Target ID — a highly specialized domain — it is essential to identify a type of QA and criteria for the references. In the absence of a benchmark for RAG in Target ID, we interviewed domain experts to inform the design of an expert-informed benchmark.

\begin{figure*}
\centering
  \includegraphics[width=2\columnwidth]{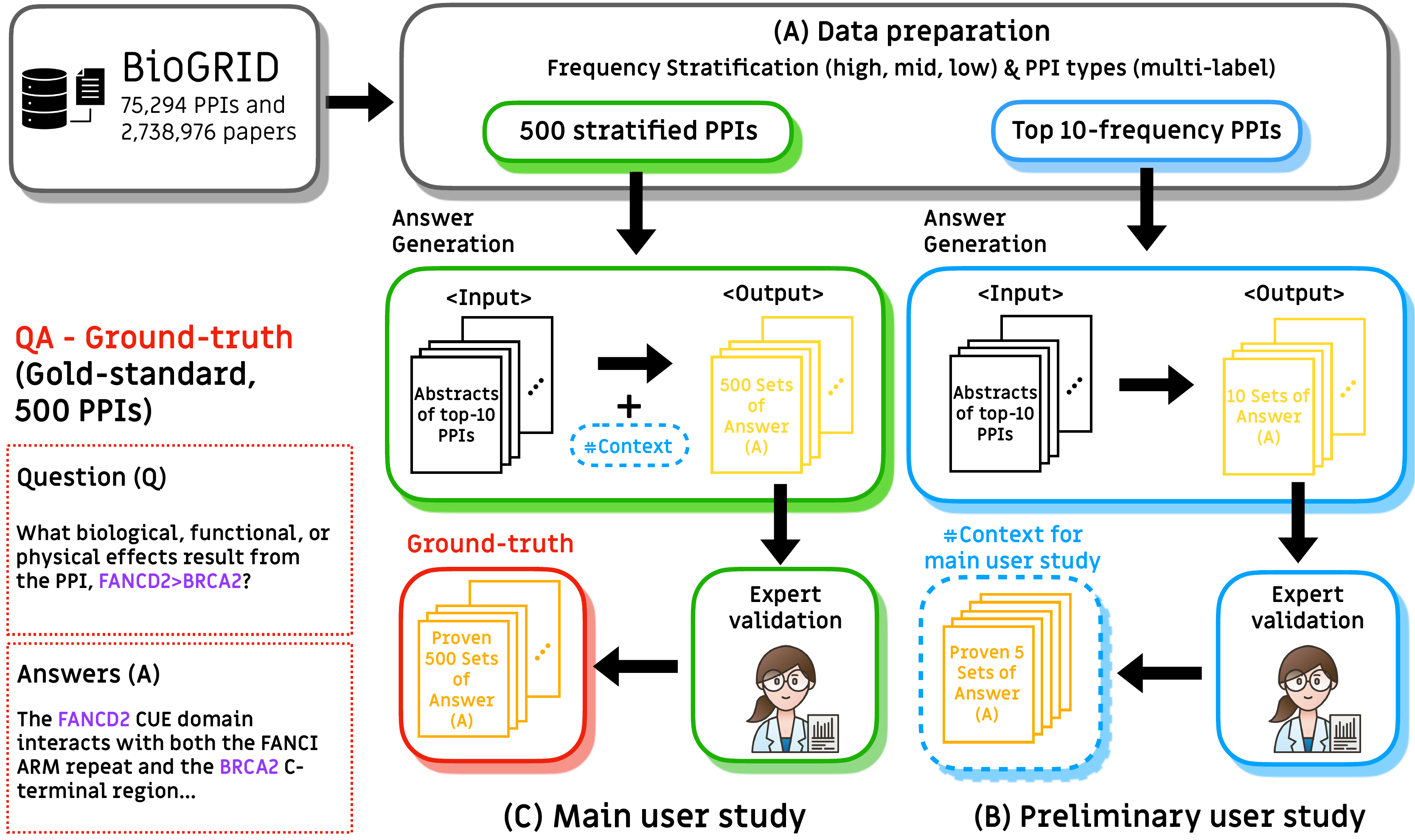}
  \caption{Overview of building the gold-standard dataset. (A) From BioGRID, we extracted the top-10 most frequent PPIs for the preliminary study and 500 stratified PPIs for the main study. (B) Preliminary study: expert validation of the top-10 PPIs provided context for answer generation in the main study. (C) Main user study: expert validation of the 500 stratified PPIs established the ground-truth QA pairs.  
  }
  ~\label{fig:dataset}
\end{figure*}

\subsection{Interview}
We interviewed five drug discovery researchers, four of whom have obtained Ph.D. degrees in chemical engineering, with two working as postdoctoral researchers at a pharmaceutical research center and two working as researchers at a pharmaceutical company. The fifth researcher is a Ph.D. candidate who has been conducting drug discovery research for the past five years in graduate school. Their work experience ranges from 7 to 10 years (mean=8.75, SD=1.3). Our findings identified 1) a type of information for QA of our benchmark, and 2) which sources to use and how to select them.


\paragraph{Type of QA} Experts prioritize biological impacts that drive therapeutic outcomes. Accordingly, we formulated a unified query: ``According to the abstract, what biological, functional, or physical effects result from {$ppi$}?'' While singular, this template is strategically designed to encapsulate the holistic rationale required for Target ID. It compels models to synthesize three core pillars: (1) interaction partners, (2) functional mechanisms (e.g., inhibition, activation), and (3) therapeutic implications. For instance,  in the Keap1–Nrf2 interaction, chemopreventive agents and oxidative stress inhibit Keap1-mediated degradation of Nrf2. As a result, Nrf2 is stabilized and activates genes with cancer-protective functions, illustrating a therapeutic mechanism against oxidative stress-related diseases.


\paragraph{Criteria for References} All experts reported using STRING\footnote{https://string-db.org/}—a database integrating multiple sources—to search for PPIs. By inputting a specific protein, users can view its associated PPIs along with descriptions based on scientific literature, which are derived from the BioGRID dataset\footnote{https://thebiogrid.org/}. 

We identified two criteria for references. First, maintaining a balanced distribution of PPI types is essential, as PPI types are informative for understanding biological processes. This balance implies high generalizability, since it allows coverage of a broader range of PPIs. Table~\ref{tab:ppi-types} in the appendix lists the seven PPI types: Reaction, Activation, Catalysis, Binding, Ptmod, Inhibition, and Expression. Second, ensuring an even distribution of PPIs across literature frequency levels is important. Frequently studied PPIs support the validation of known mechanisms, whereas infrequently studied ones are often the focus of novel exploration. Thus, literature frequency serves as a strong indicator of potential utility in these workflows.

\begin{table}[t]
    \centering
    \caption{Distribution of frequency levels for \system. The Gold and Silver datasets are balanced across literature frequency.}
    \label{tab:dataset_frequency}
    \renewcommand{\arraystretch}{1.1} 
    \resizebox{\columnwidth}{!}{%
        \begin{tabular}{lccc}
            \toprule
            \textbf{Frequency} & \textbf{Gold (500)} & \textbf{Silver (3,720)} & \textbf{Total (4,220)} \\
            \midrule
            High & 166 (33.2\%) & 1,275 (34.3\%) & 1,441 (34.1\%) \\
            Medium & 167 (33.4\%) & 1,246 (33.5\%) & 1,413 (33.5\%) \\
            Low & 167 (33.4\%) & 1,199 (32.2\%) & 1,366 (32.4\%) \\
            \midrule
            \textbf{Total} & 500 (100.0\%) & 3,720 (100.0\%) & 4,220 (100.0\%) \\
            \bottomrule
        \end{tabular}%
    }
\end{table}

\subsection{Task}
Based on the insight from the interviews, we designed an End-to-End RAG task that evaluates how well the output of the RAG model supports the ground-truth biological process. The task is to generate an answer (A) to a question (Q) concerning the biological, functional, or physical effects of protein-protein interactions (PPIs) that contribute to a therapeutic impact, using evidence derived from BioGRID-curated paper abstracts. These abstracts are selected to ensure a balanced distribution of PPI types and frequencies.

\section{Dataset}
We constructed a benchmark dataset through three main steps:
(1) a gold-standard dataset consisting of 500 expert-verified PPIs collected via a user study;
(2) an auto-evaluation model, implemented as an ensemble of LLMs reflecting expert labeling features; and
(3) a silver-standard dataset of 3,720 PPIs automatically labeled using an auto-evaluation model.
\footnote{\system will be released under a CC BY 4.0 license and made publicly available upon publication.}

\subsection{Gold-standard dataset}
We constructed an expert-verified dataset focusing on a PPI through two main phases: 1) data preparation with BioGRID datasets by PPI type and literature frequency, and 2) expert annotation through user studies.

\subsubsection{Data Preparation}\label{data_preparation}
In the data preparation phase (Figure~\ref{fig:dataset}-A), we conducted frequency stratification and PPI types from the BioGRID dataset~\cite{oughtred2021biogrid} including 2,738,976 PPI samples from 75,294 papers. 
Stratification prevented the dataset from being overloaded with common diseases (e.g., COVID-19, cancer) and ensured broader coverage, including diabetes, mental disorders, and neurodegenerative diseases.

For frequency stratification, PPIs were categorized into high-, mid-, and low-frequency groups based on their occurrence in the literature. Interactions with a frequency of 1 were labeled as \textit{Low}, while those with frequency $\geq 2$ were further divided by computing the mean frequency $\mu$: interactions with frequency $\geq \mu$ were labeled as \textit{High}, and those with frequency $< \mu$ as \textit{Medium}. We extracted 8,066 papers that address only single PPIs to ensure that the benchmark reflects the sufficient retrieval context required for RAG-based QA.

For PPI types, we assigned multi-label annotations to each PPI from seven predefined interaction types—for instance, a kinase might both bind to and phosphorylate its substrate (Binding and Ptmod). After multi-label classification, a dataset with 500 PPIs is created for the following expert's evaluation with equal sampling from each group to mitigate frequency bias while preserving a balanced distribution across the high-, mid, and low-frequency categories (Table~\ref{tab:dataset_frequency}) and PPI type categories (Table \ref{tab:dataset_ppitype} in the appendix).
Through these steps, we prepared the top-10 PPIs for the preliminary user study and 500 stratified PPIs for the user study, which are mutually exclusive (Figure~\ref{fig:dataset}-A).





\begin{figure*}
\centering
  \includegraphics[width=2
  \columnwidth]{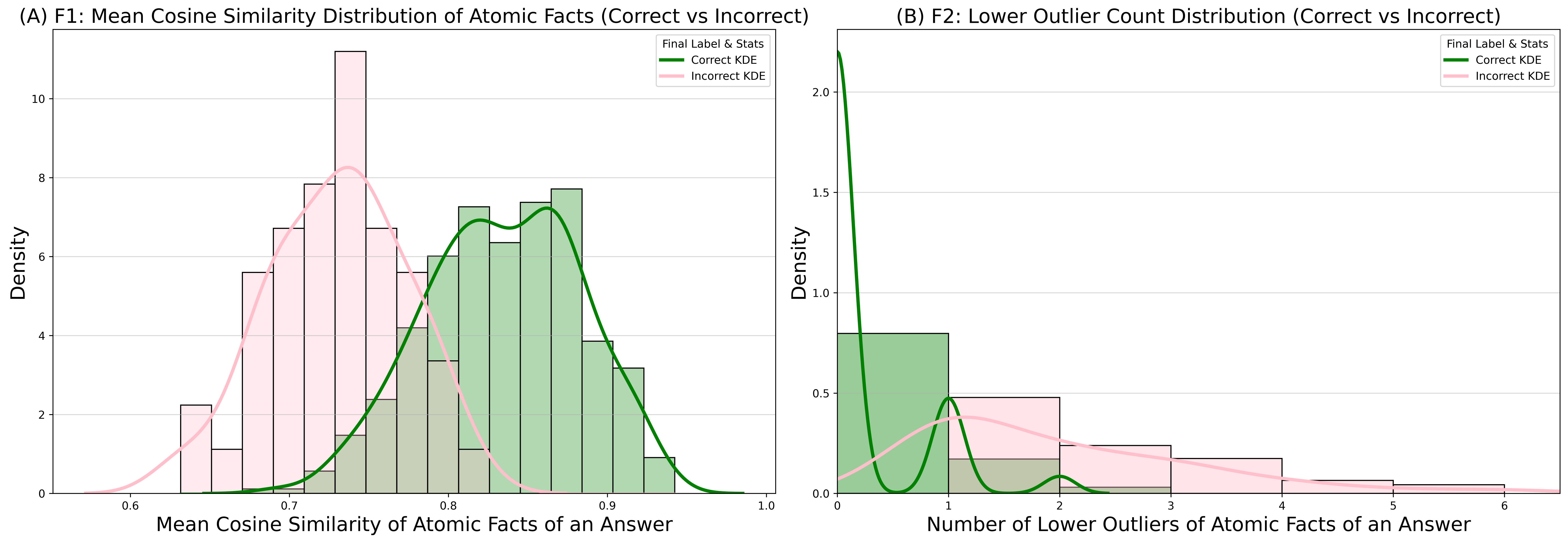}
  \caption{Distributions of semantic features ($F_1$ and $F_2$) of Correct (green) and Incorrect (pink) answers based on user study labeling: (A) Distribution of the mean cosine similarity between atomic facts of an answer and its corresponding abstract ($F_1$); (B) Distribution of the number of atomic facts considered as lower outliers ($F_2$). Correct answers show higher similarity and fewer outliers compared to Incorrect answers.}
  ~\label{fig:fig_sim}
\end{figure*}


\subsubsection{Expert Labeling}
We conducted preliminary and main user studies in which domain experts validated LLM-generated answers, resulting in a gold-standard dataset of 500 expert-validated PPIs.



\textbf{Step 1: Preliminary study.} 
Figure~\ref{fig:dataset}-B shows the whole process of the preliminary study. 
We provided GPT-4o with the abstracts of the top-10 PPIs and the template (Table~\ref{tab:Prompt} in the appendix) to generate answers, which were evaluated by three experts as perfect, acceptable, or inaccurate.
\begin{itemize}
    \item Perfect: The response correctly answers the user’s question and contains no hallucinated content.
    \item Acceptable: The response provides a useful answer but may contain minor inaccuracies that do not significantly reduce its utility.
    \item Incorrect: The response contains incorrect or irrelevant information that fails to address the user’s question.
\end{itemize}
We selected 5 PPIs that were rated as perfect by all experts as the final context examples for the LLM in the main user study. 
Details are provided in {\S}~\ref{preliminary} in the appendix.


\begin{table}[t]
  \centering
  \caption{Distribution of initial and final labels for the 500 PPI dataset. Perfect and Acceptable were merged into the Correct category for final labeling.}
  \label{tab:labeling_500}
  \renewcommand{\arraystretch}{1.3}
  \resizebox{\columnwidth}{!}{
  \begin{tabular}{c|c|c|c}
        \toprule
        \textbf{Initial label} & \textbf{\# (\%)} & \textbf{Final label} & \textbf{\# (\%)} \\
        \midrule
        Perfect & 383 (76.6\%) & \multirow{2}{*}{Correct} & \multirow{2}{*}{454 (90.8\%)} \\
        Acceptable & 71 (14.2\%) & & \\
        \midrule
        Incorrect & 46 (9.2\%) & Incorrect & 46 (9.2\%) \\
        \midrule
        \textbf{Total} & \textbf{500} & \textbf{Total} & \textbf{500} \\
        \bottomrule
  \end{tabular}
  }
  \renewcommand{\arraystretch}{1}
\end{table}

\textbf{Step 2: Main user study.} 
Figure~\ref{fig:dataset}-C shows the whole process of the main user study. 
We provided GPT-4o with the abstracts of the 500 stratified PPIs, the template, and five abstract–answer PPI pairs selected from the preliminary study as context for answer generation.
The answers were then evaluated by ten experts as perfect, acceptable, or inaccurate.
Based on their review, expert-verified labels were collected, resulting in a ground-truth QA dataset. When answers were labeled as acceptable or inaccurate, the experts were asked to revise them into perfect answers. Details are provided in {\S}~\ref{main_study} in the appendix.

\textbf{Step 3: Gold-standard dataset.}  Table~\ref{tab:labeling_500} shows the results of expert labeling. The expert-labeled distribution was 383 (76.6\%) Perfect, 71 (14.2\%) Acceptable, and 46 (9.2\%) Inaccurate. We merged the labels, Perfect and Acceptable, into \textbf{Correct}, resulting in two final labels, \textbf{Correct} and \textbf{Incorrect}. 
In addition, the number of samples in which experts revised more than a clause is 482 (96.4\%). This intensive expert intervention ensures that the final dataset reflects verified expert knowledge rather than model biases. Finally, we built a gold-standard dataset consisting of 500 QA pairs, each paired with the question and both the corresponding ground-truth and labels.


\subsection{Auto-evaluation model}
We propose an ensemble-based auto-evaluation model that integrates three sub-LLMs, which are specialized experts' labeling features that distinguish accurate from inaccurate responses: the distribution of average atomic fact-level similarities between each answer and the abstract ($F_1$), and the distribution of low-similarity atomic facts ($F_2$). Based on statistical analysis, we identify the expert labeling features and reflect them on each sub-model.

\begin{table}[t]
\caption{Results on error rates (ER). The ensemble model achieves the lowest total ER (bold) compared to experts.}
  \label{tab:evaluator_comparison}
    \centering
    \renewcommand{\arraystretch}{1.1}
    \resizebox{\columnwidth}{!}{%
    \begin{tabular}{lccc}
        \toprule
        \multirow{2}{*}{\textbf{Evaluator}} & \multicolumn{3}{c}{\textbf{Error Rate (\%)}} \\
        \cmidrule(lr){2-4}
         & \textbf{Acc (35)} & \textbf{Inacc (35)} & \textbf{Total (70)} \\
        \midrule
        Expert & 0.00 & 0.00 & 0.00 \\
        \midrule
        Only GPT-4o & 31.43 & 74.29 & 52.86 \\
        GPT-4o (w/ Abs) & 5.71 & 100.00 & 53.00 \\
        GPT-4o (w/ Abs+GT) & 0.00 & 83.00 & 42.00 \\
        \midrule
        M1 (GPT-4o-based) & 8.57 & 8.57 & 8.57 \\
        M2 (GPT-4o-based) & 14.29 & 2.86 & 8.57 \\
        M3 (GPT-4o-based) & 2.86 & 74.29 & 38.57 \\
        \midrule
        \textbf{Ensemble (M1--3)} & \textbf{2.86} & \textbf{5.71} & \textbf{4.29} \\
        \bottomrule
    \end{tabular}%
    }
\end{table}

\subsubsection{Expert Labeling Features}
We characterized the two features ($F_1$ and $F_2$) of correct and incorrect answers by measuring how their atomic facts semantically align with the corresponding abstract. $F_1$ represents the distribution of average atomic fact-level similarities between each answer and the abstract. We decomposed answers and abstracts into atomic facts~\cite{min2023factscore} by using GPT-4o. For each answer, we compute a cosine similarity of each atomic fact of an answer to each atomic fact of the abstract, and then define the maximum value as the similarity for an atomic fact of an answer. $F_1$ is the collection of these per-answer averages. $F_2$ means the distribution of low-similarity atomic facts. For each answer, we count the number of atomic facts whose similarity falls below a threshold (mean minus two standard deviations of $F_1$, 0.61). $F_2$ is defined as the collection of these counts. 
The pseudo algorithm for $F_1$ and $F_2$ is shown in Algorithm \ref{alg:similarity_features}.

An independent t-test was conducted to reveal that Correct exhibits higher cosine similarity and fewer low-similarity outliers compared to Incorrect (Figure~\ref{fig:fig_sim})
In the case of $F_1$ (Figure~\ref{fig:fig_sim}-A), we observed a significant difference in the cosine similarity of atomic facts of answers to abstract between Correct and Inaccurate (\textit{p} < 0.001). The Correct group had a mean of 0.83 (SD = 0.04), while the Incorrect group had a mean of 0.72 (SD = 0.04). In the case of $F_2$ (Figure~\ref{fig:fig_sim}-B), we observed a significant difference in the number of low-similarity outliers of atomic facts of answers to abstract between Correct and Inaccurate (\textit{p} < 0.001). The Correct group had a mean of 0.22 (SD = 0.48), while the Incorrect group had a mean of 1.37 (SD = 1.27).

\subsubsection{Ensemble LLM for auto-evaluation}
We propose an ensemble auto-evaluation model composed of three LLMs (GPT-4o). Two of the sub-LLMs assess factual correctness at the atomic level ($F_1$ and $F_2$), while the third evaluates global semantic alignment with the ground truth (GT). We used OpenAI’s text-embedding-3-small model\footnote{\url{https://openai.com/index/new-embedding-models-and-api-updates/}}. for embeddings to compute the cosine similarity.

\begin{itemize}
  \item Model-1 (M1): GPT-4o evaluates the answer based on the average cosine similarity between its atomic facts and the corresponding abstract by comparing it with the distribution of GT (Correct: 0.83~$\pm$~0.04; Incorrect: 0.72~$\pm$~0.04).
  \item Model-2 (M2): GPT-4o judges the answer by counting the number of atomic facts that are considered lower outliers in their cosine similarity to the abstract and comparing it with the distribution of GT (Correct: 0.22~$\pm$~0.08; Incorrect: 1.37~$\pm$~1.27).
  \item Model-3 (M3): GPT-4o assesses how well the answer is semantically supported by the ground-truth answer (GT) as a whole.
\end{itemize}

For each feature, we selected a total of 18 representative examples—three labeled as accurate and three as inaccurate for each model—to serve as demonstrations within the prompt.
Each sub-LLM predicts a label as an output. Based on the predictions from the three sub-LLMs, the final label is determined through majority voting.

Table~\ref{tab:evaluator_comparison} shows the better performance of our ensemble model compared to other models in evaluating responses. To ensure a balanced class distribution in the evaluation, we randomly selected 35 samples from the Correct class and 35 samples from the Incorrect class, excluding those used as examples for the models. In total, 70 samples were used for evaluation. Models prompted with expert-labeled features exhibit more balanced performance across both Correct and Incorrect classes compared to models without such prompts. In contrast, general (non-prompted) models tend to overpredict the Correct label in most cases, highlighting a discrepancy from expert knowledge. Such bias may indicate a potential starting point for hallucination issues in domain-specific tasks.
To better understand these discrepancies, we present representative failure cases (Figures~\ref{fig:incorrect_example} and~\ref{fig:false_pos_example} in the appendix) reveal distinct error patterns: false positives driven by outcome bias despite missing mechanisms, and false negatives caused by lexical rigidity, underscoring the necessity of our ensemble approach to balance atomic verification with holistic reasoning.

\subsection{Silver-standard dataset}
We applied our ensemble auto-eval LLM in generating GT for a silver-standard dataset consisting of 3,720 PPIs. We generated answers for 5,000 randomly selected samples from a total of 8,066 using the same GPT-4o model and prompts employed in constructing the gold-standard dataset ({\S}~\ref{data_preparation}).
By using M1 and M2 of our ensemble auto-eval LLM, we labeled an AI-generated answer as GT if both M1 and M2 evaluated it as Accurate. 
Notably, we excluded M3 from this process to prioritize factual grounding over stylistic similarity. M3 measures alignment with the reference text; thus, its exclusion prevents the propagation of stylistic biases and ensures that the silver-standard labels remain strictly evidence-based. Only those answers were included in the final silver-standard dataset.
The pseudo algorithm for auto-evaluation is shown in Algorithm \ref{alg:auto_eval_ensemble}. Consequently, we built \system, a factual question answering benchmark of 4,420 question-answer pairs (Table~\ref{tab:dataset_frequency}).



\begin{algorithm}
  \small
\caption{{Computing $F_1$ (average similarities) and $F_2$ (low similarity counts)}}
\label{alg:similarity_features}
\begin{algorithmic}[1]
\Require Answer set $\{A_1, A_2, \dots, A_n\}$, Abstract set $\{B_1, B_2, \dots, B_k\}$
\Ensure Features $F_1$ and $F_2$
    \State \textbf{Input}: Answer set $\{A_1, A_2, \dots, A_n\}$, Abstract set $\{B_1, B_2, \dots, B_k\}$
    \State Extract atomic facts from Abstract set: $B = \{b_1, b_2, \dots, b_l\}$
    \State Initialize list $S$ to store per-answer average similarities
    \State Initialize list of lists $\{S_1, S_2, \dots, S_n\}$ for per-answer fact similarities
    \For{each Answer $A_i \in \{A_1, A_2, \dots, A_n\}$}
        \State Extract atomic facts from $A_i$: $A_i^{\text{facts}} = \{a_1, a_2, \dots, a_{m_i}\}$
        \State Initialize list $S_i$ to store representative similarities for $A_i$
        \For{each atomic fact $a_j \in A_i^{\text{facts}}$}
            \State Compute representative similarity: $s_j \gets \max_{k \in \{1, \dots, l\}} \text{sim}(a_j, b_k)$
            \State Add $s_j$ to $S_i$
        \EndFor
        \State Compute average similarity for $A_i$: $\bar{S}_i \gets \frac{1}{|S_i|} \sum_{s \in S_i} s$
        \State Add $\bar{S}_i$ to $S$
        \State Store $S_i$ for later use
    \EndFor
    \State Set $F_1 \gets S$ \Comment{$F_1$ is the distribution of average similarities}
    \State Compute threshold $T \gets \text{mean}(S) - 2 \times \text{std}(S)$
    \State Initialize list $L$ to store low similarity counts per answer
    \For{each $S_i$ in $\{S_1, S_2, \dots, S_n\}$}
        \State Compute $L_i \gets$ number of $s_j \in S_i$ where $s_j < T$
        \State Add $L_i$ to $L$
    \EndFor
    \State Set $F_2 \gets L$ \Comment{$F_2$ is defined as the distribution of low similarity counts per answer}
    \State \textbf{Output}: Feature $F_1$ (distribution), Feature $F_2$ (low similarity counts)
\end{algorithmic}
\end{algorithm}
\begin{algorithm}
\small
\caption{{Auto-eval ensemble decision rule for constructing the silver-standard dataset}}
\label{alg:auto_eval_ensemble}
\begin{algorithmic}[1]
\Require Labels from Model-1 and Model-2 for answer $A$
\Ensure Final decision: ``GT'' or ``Not GT''
\State \textbf{Input}: Model-1 label $L_1$, Model-2 label $L_2$
\If{$L_1 = \text{accurate}$ \textbf{and} $L_2 = \text{accurate}$}
    \State \Return ``GT'' \Comment{Add to silver-standard dataset}
\Else
    \State \Return ``Not GT''
\EndIf
\end{algorithmic}
\end{algorithm}

\section{Benchmarking Experiments}
In this section, we present the performance of LLMs and RAG systems on \system, demonstrating that \system has a reasonable level of difficulty and applicability.


\begin{table*}
\caption{Performance of LLM- and RAG-based models (default and our paper corpus) in terms of atomic fact similarity ($F_1$), low-similarity fact counts ($F_2$), and accuracy. The values of $F_1$ and $F_2$ are shown as mean and standard deviation. \textbf{Bold font} represents the best methods in the evaluation. } 
  \label{tab:benchmarking_performance}
    \centering
    {
    \fontsize{8pt}{10pt}\selectfont
    \renewcommand{\arraystretch}{1.5}
    \begin{tabular}{r r|p{1.5cm}<{\centering}|p{1.5cm}<{\centering}|cccc}
        \toprule
        \multicolumn{2}{c|}{\multirow{3}{*}{\textbf{Model}}} & \multicolumn{2}{c|}{\textbf{Cosine similarity}} & \multicolumn{4}{c}{\multirow{2}{*}{\textbf{Accuracy}}} \\
        & & \multicolumn{2}{c|}{\textbf{of atomic facts}} & & & & \\
        & & \multicolumn{1}{c}{\textbf{$F_1$}} & \textbf{$F_2$} & \textbf{M1} & \textbf{M2} & \textbf{M3} & \textbf{Ensemble}\\
        \cline{1-8}
        \multirow{3}{*}{LLM-based} & GPT-4.1 & 0.72~$\pm$~0.07 & 3.09~$\pm$~3.54 & 52.69\% & 29.30\% &36.83\% & 39.52\%\\
        & Gemini 2.0 flash & 0.74~$\pm$~0.13 & \textbf{0.67~$\pm$~1.53} & 58.87\% & \textbf{67.74\%} & 11.02\% & 56.18\%\\
        & MedLlama & 0.68~$\pm$~0.11 & 3.39~$\pm$~3.42 & 39.25\% & 25.27\% & 15.86\% & 25.27\%\\
        \cline{1-8}
        \multirow{2}{*}{\shortstack{General RAG-based \\ (our corpus)}} & Vanila + Re-rank & 0.64~$\pm$~0.15 & 2.76~$\pm$~3.86 & 57.26\% & 66.94\% & 61.29\% & 59.95\%\\
        & GraphRAG & 0.63~$\pm$~0.12 & 4.95~$\pm$~4.01 & 42.47\% & 47.85\% & 44.35\% & 45.97\%\\
        \cline{1-8}
        \multirow{2}{*}{\shortstack{Bio RAG-based \\ (default corpus)}} & MedRAG & 0.56~$\pm$~0.10 & 7.53~$\pm$~4.25 & 8.58\% & 6.17\% & 20.38\% & 8.04\%\\
        & GraPPI & 0.51~$\pm$~0.20 & 2.39~$\pm$~3.51 & 17.20\% & 17.74\% & 5.11\% & 13.44\%\\
        & GeneGPT & 0.70~$\pm$~0.07 & 3.68~$\pm$~3.79 & 45.16\% & 22.85\% & 47.31\% & 35.48\%\\
        \cline{1-8}
        \multirow{2}{*}{\shortstack{Bio RAG-based \\ (our corpus)}} & MedRAG & 0.71~$\pm$~0.15 & 3.36~$\pm$~4.46 & 55.76\% & 59.79\% & 55.50\% & 55.23\%\\
        & GraPPI & \textbf{0.76~$\pm$~0.09} & 2.46~$\pm$~4.02 & \textbf{74.26\%} & 64.61\% & \textbf{78.28\% }& \textbf{74.80\%}\\

        \bottomrule
    \end{tabular}
    }
\end{table*}

\paragraph{Experiment setup}  We tested eight systems including two general closed-source LMs: ChatGPT-4.1(gpt-4.1) \cite{openai2025gpt41} and Gemini-2.0-Flash(gemini-2.0-flash) \cite{google2025gemini2flash}; one open-source LM fine-tuned for the biomedical QA task: JSL-MedLlama-3-8B-v2.0~\cite{johnsnowlabs2024medllama}; general RAGs: Vanilla RAG with Re-ranking~\cite{lewis2020rag, nogueira2019reranking} and GraphRAG~\cite{edge2024graphrag}; and
three biomedical RAG systems: GraPPI~\cite{li2025grappi}, MedRAG~\cite{xiong2024benchmarking}, and GeneGPT \cite{jin2024genegpt}.


We apply the chat template to the open-source model and run inferences with a single A100 GPU. We use Gemini-2.0-flash for the base model in GraPPI and GPT-4.1 for MedRAG. 
To investigate the quantitative relationship between the quality of retrieval results and overall performance, we compared RAG-based models using their default corpus versus our paper corpus for retrieval.
As GeneGPT does not support corpus substitution, it was excluded from the paper-corpus setting. 
The prompt consists of instructions and five QA pairs as few-shot examples, which are validated by experts (Table~\ref{tab:Prompt} in the appendix). 
We sample 372 PPIs from our database, preserving the distribution of frequency levels and PPI types, and generate biological, functional, or physical effects from each PPI. For each system, a QA pair $\mathcal{P}(Q_{ppi}, A_{sys})$ would be generated. Once the results are generated, the question-answer pairs, $\mathcal{P}(Q_{ppi}, A_{sys1}, A_{sys2},....,A_{sys6})$, are passed to our auto-eval LLM.


\paragraph{Results} Table~\ref{tab:benchmarking_performance} shows performance of the LLM- and RAG-based models on the cosine similarity of the atomic facts with the reference ($F_1$), low-similarity fact counts ($F_2$), and the accuracy. 
Overall, RAG models using our paper corpus outperformed both LLM-based models and RAG models with default corpora. GraPPI with our paper corpus achieved the best results, with the highest atomic fact similarity ($F_1 = 0.76$) and ensemble accuracy (74.80\%). Given that GraPPI is specialized for protein–protein interaction tasks~\cite{li2025grappi}, this result indicates how domain-specific corpora and domain-specific models complement each other in improving RAG performance. 

Interestingly, LLM-based models, despite the absence of retrieval, consistently outperformed RAG models with default corpora. 
Their higher $F_1$ and $F_2$ scores indicate that LLMs generally produce answers aligned with the ground truth, underscoring that an ill-suited corpus can substantially undermine performance.
By capturing such contrasts between corpus and model effects, our benchmark effectively evaluates both retrieval and generation, the core processes of RAG.
At the same time, LLMs tended to produce generally correct but incomplete answers. 
For example, Gemini achieves high accuracy in M1 (58.87\%) and M2 (67.74\%) but low accuracy in M3 (11.02\%).  
While LLMs capture broad factual accuracy, they fail to cover the full scope of domain-specific content.
This underscores the importance of ensemble-based approaches that consider both answer-level accuracy (M1 and M2) and alignment with ground-truth that experts validate (M3).

To explicitly assess how retrieval quality affects end-to-end performance, we conducted comparative experiments where RAG-based models retrieved evidence from either their default corpus or our curated paper corpus. The results show performance improvements when using our paper corpus, indicating that the benchmark is indeed sensitive to retrieval quality and not only to the generation ability of the underlying LLM. For example, GraPPI’s accuracy increased from 13.44\% to 74.80\%, and MedGPT’s accuracy from 8.04\% to 55.23\% when switching from the default corpus to our paper corpus. These patterns confirm that retrieval quality is a primary driver of performance differences, demonstrating that our benchmark meaningfully evaluates the retrieval component of RAG pipelines.

\section{Conclusion}
This paper proposes \system, an expert-validated benchmark featuring 4,220 QA pairs designed to advance retrieval-augmented generation (RAG) research in target identification.
\system utilizes a unified query template derived from expert studies that reflects a complex, multi-step reasoning chain: (1) Entity Identification, (2) Mechanism Elucidation, and (3) Therapeutic Impact. To ensure both high quality and scale, we first established a Gold dataset of 500 expert-curated QA pairs, which was then expanded into a 3,720-pair Silver dataset using an ensemble LLM-based auto-evaluation framework.
Empirical studies with existing LLMs and RAG models demonstrate \system's applicability and provide valuable insights for future improvements.
This ensures that \system remains at the forefront of RAG research, driving progress in both robust NLP modeling and real-world biomedical applications.
We plan to further enhance \system by diversifying question types to include multi-hop and comparative reasoning tasks, and extending its scope to broader scientific discovery challenges.

\section{Limitation}


We present an RAG benchmark for Target ID and an auto-evaluation model, but our study still has some limitations. 
First, although we built our gold-standard dataset of 500 QA pairs, the number of samples may still be insufficient to cover the full range of biological impacts of PPIs. 
Second, while we identified the two features with ten experts, their perspectives may not fully capture the expertise of the drug discovery domain. 
Third, the current benchmark primarily focuses on a single question type regarding direct biological impacts, which may limit the evaluation of diverse reasoning patterns, such as multi-hop, required in broader NLP tasks. 
Fourth, our dataset represents a static snapshot of biomedical knowledge and does not yet account for the temporal evolution of scientific findings. There is a potential risk that misuse of our benchmark or evaluation model outside expert-supervised research contexts could lead to misleading conclusions in therapeutic discovery. 
Lastly, relying solely on abstracts may overlook fine-grained biological details and figures present in full-text articles.

As future work, we plan to implement a rigorous maintenance protocol using Semantic Versioning (e.g., v1.0, v2.0) aligned with STRING dataset updates to track temporal shifts, and continue collecting annotated data through collaboration with additional experts to enhance the generality of \system by identifying more tasks and integrating ground-truth.

\section{Ethical Considerations}
This work uses publicly available biomedical literature and protein–protein interaction data and does not include personally identifying information or sensitive personal data. \system is released as a research benchmark to support studies on retrieval-augmented generation for drug discovery and is not intended for direct clinical or commercial deployment. AI-assisted tools were used to support language editing and clarity of the manuscript; all scientific content and conclusions were produced and verified by the authors.

\bibliography{anthology,custom}

\clearpage
\appendix
\begin{table*}[h!]
\centering
\resizebox{\textwidth}{!}{
\begin{tabular}{l|c|c|c|c|c}
\textbf{Model} & \textbf{ROUGE-1 F1} & \textbf{ROUGE-L F1} & \textbf{BERT Precision} & \textbf{BERT Recall} & \textbf{BERT F1} \\
\hline
gpt-4o & 0.4495 & 0.2273 & 0.6625 & 0.6041 & \textbf{0.6314} \\
gemini & 0.3284 & 0.2200 & 0.6292 & 0.5369 & 0.5788 \\
grok-2-1212 & 0.4468 & \textbf{0.2342} & \textbf{0.6639} & 0.5980 & 0.6289 \\
deepseek-v3 & \textbf{0.4673} & 0.2189 & 0.5628 & \textbf{0.6158} & 0.5878 \\
\end{tabular}
}
\caption{Comparison of models on ROUGE and BERT-based evaluation metrics. Bolded values indicate best performance per column.}
\label{tab:results}
\end{table*}

\section{Appendix}

\subsection{Type of PPIs and the distribution}
Through expert interviews, we identified seven types of PPIs considered in their workflows (Table~\ref{tab:ppi-types}). Based on this insight, we extracted 500 PPIs with a balanced distribution across these types (Table~\ref{tab:results}).

\begin{table*}
  \caption{Descriptions and examples of seven PPI types~\cite{szklarczyk2021string, szklarczyk2023string}}
  \label{tab:ppi-types}
  \centering
    \begin{tabular}{p{2cm} p{11cm}}
    \toprule
    \multicolumn{1}{c}{\textbf{PPI Type}} & 
    \multicolumn{1}{c}{\textbf{Description}} 
    \\
    \midrule
    Reaction   & Proteins participate in the same biochemical reaction or metabolic pathway step, often linked as successive enzymes. \\
    Activation & One protein increases the activity of another, often by direct interaction or signaling. \\
    Catalysis  & One protein enzymatically modifies another, such as a kinase phosphorylating a substrate. \\
    Binding    & TProteins physically interact to form a complex, either stable or transient. \\
    Ptmod      & One protein chemically modifies another after translation.\\
    Inhibition & One protein decreases or suppresses the activity of another protein. \\
    Expression & One protein regulates the expression level (amount produced) of another, typically at the transcriptional level. \\
    \bottomrule
  \end{tabular}
\end{table*}

\begin{table*}[]  
\caption{Distribution of PPI interaction types for \system: the gold-standard and silver-standard datasets.
}
  \label{tab:dataset_ppitype}
    \centering
    \renewcommand{\arraystretch}{1.6}
    \begin{tabular}{r|c|c|c}
        \toprule
        \multicolumn{1}{c|}{\textbf{PPI Types}} & \textbf{Gold-standard (500)} & \textbf{Silver-standard (3,720)} & \textbf{Total (4,220)} \\
        \cline{1-4}
        Reaction & 85 (17.0\%) & 205 (5.5\%) & 290 (6.9\%) \\
        Activation & 107 (21.4\%) & 937 (25.2\%) & 1044 (24.7\%) \\
        Catalysis & 86 (17.2\%) & 231 (6.2\%) & 317 (7.5\%) \\
        Binding & 428 (85.6\%) & 3273 (88.0\%) & 3701 (87.7\%) \\
        Ptmod & 122 (24.4\%) & 1470 (39.5\%) & 1592 (37.7\%) \\
        Inhibition & 103 (20.6\%) & 1108 (29.8\%) & 1211 (28.7\%) \\
        Expression & 152 (30.4\%) & 1470 (39.5\%) & 1622 (38.4\%) \\
        \cline{1-4}
        \textbf{Total} & 500 (100.0\%) & 3720 (100.0\%) & 4220 (100.0\%) \\
        \bottomrule
    \end{tabular}
\end{table*}

\subsection{Preliminary user study}\label{preliminary}
To ensure the quality of answers for the main user study, we conducted a preliminary user study. All participants were researching Target ID. Their career experience ranged from 4 to 9 years (mean=6, SD=2.1). Our study was approved by the Institutional Review Board (IRB), and the consent of the participants was sought before the study. Each participant received a \$30 gift certificate. 

\begin{figure*}
  \centering
  \includegraphics[width=1\columnwidth]{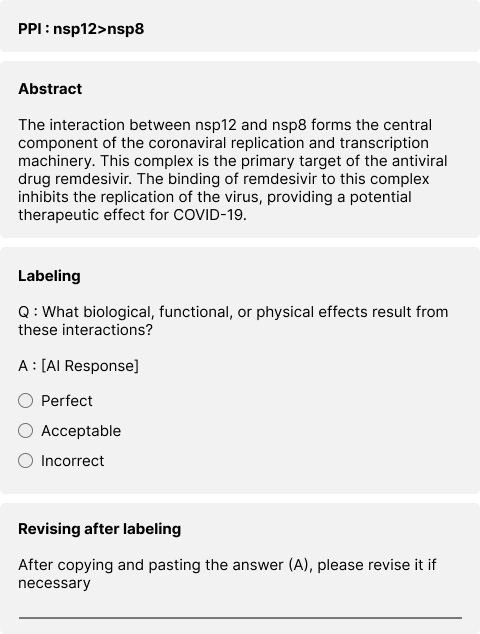}
  \caption{The example of the questionnaire used in our main user study, where experts evaluated an AI-generated answer of the biological or functional effects of PPIs based on the corresponding abstract, and revised them if necessary.}
  \label{fig:google_form}
\end{figure*}

\begin{figure*}[h!]
    \centering
    \begin{minipage} {1\textwidth}
        \centering
        \includegraphics[width=1\linewidth]{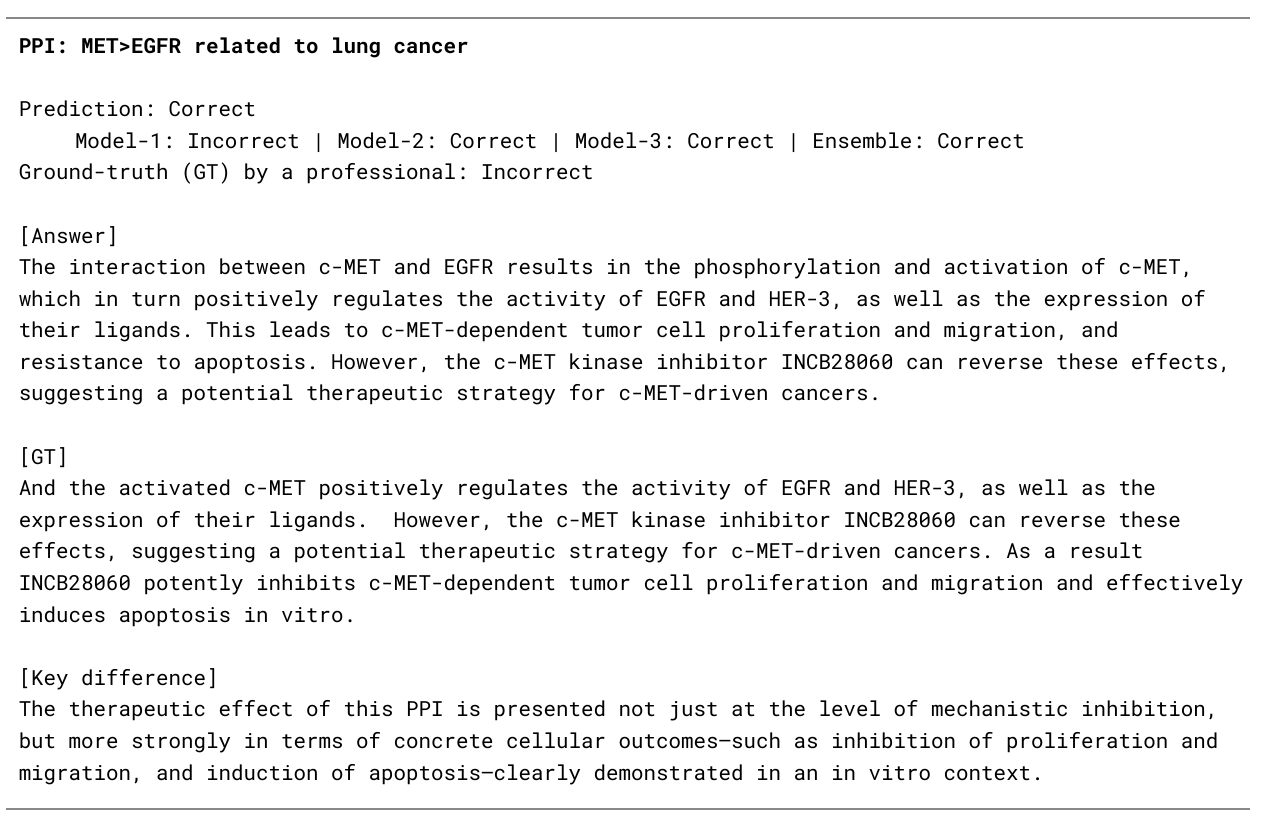}
        \caption{A false positive case where the model evaluated the answer as correct based on general mechanistic descriptions, whereas experts judged it incorrect due to the absence of specific experimental evidence (e.g., in vitro outcomes) required for therapeutic validation..}
        \label{fig:incorrect_example}
    \end{minipage}

    \begin{minipage}{1\textwidth}
         \centering
        \includegraphics[width=1\linewidth]{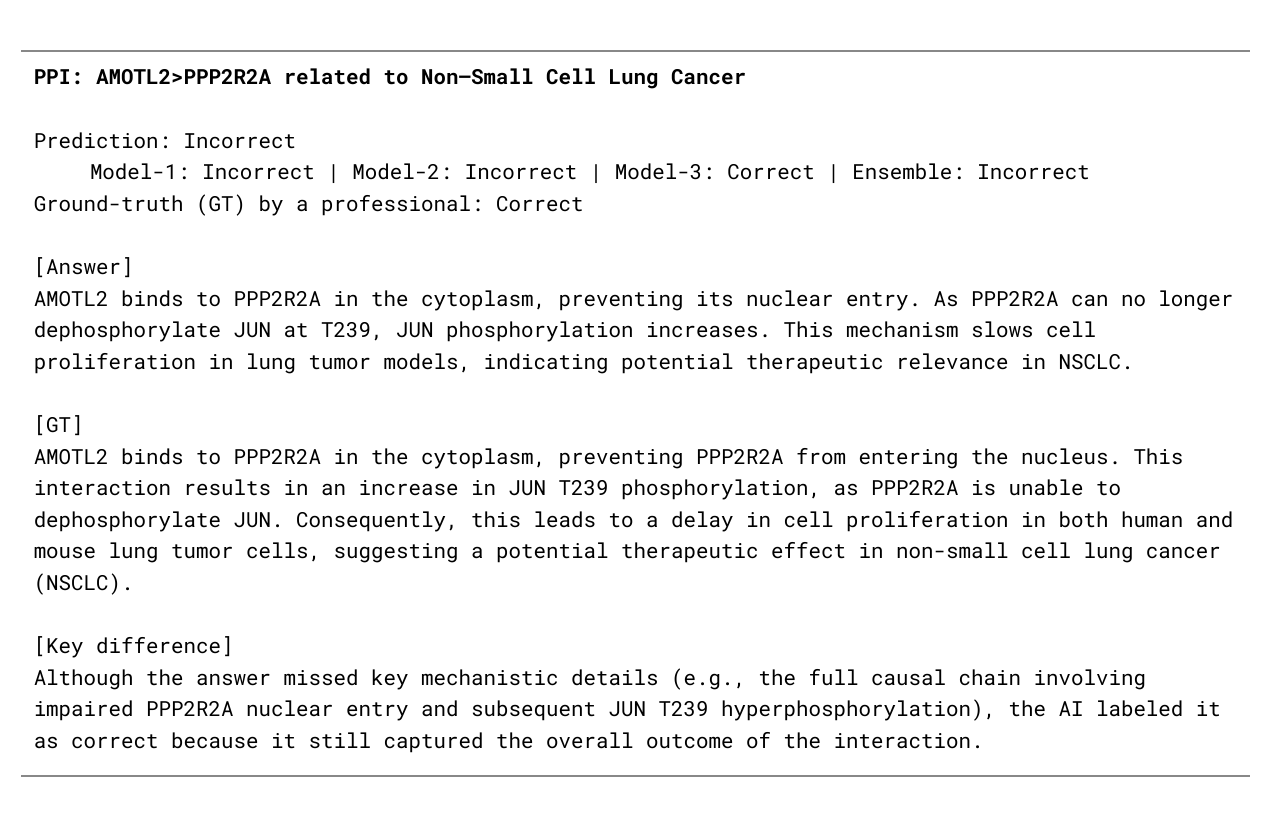}
        \caption{A false negative case where the model incorrectly penalized a semantically accurate answer due to lexical variations (e.g., preventing its nuclear entry vs. preventing PPP2R2A from entering the nucleus), highlighting the challenge of automated evaluators in recognizing synonymous phrasing..}
        \label{fig:false_pos_example}
    \end{minipage}
\end{figure*}



To ensure the generalizability of the context, we selected the top 10 PPIs with the highest frequency. We used GPT-4o in generating answers for this study based on the performance ({\S}~\ref{model-selection}). We feed a prompt containing the abstract of each of the 10 selected PPIs along with the following the question to GTP-4o, and obtain the corresponding answer:
\begin{itemize}
  \item Question: What biological, functional, or physical effects result from \textit{<PPI>}?
\end{itemize}

Experts were asked to label 10 sets of answers as Perfect, Acceptable, or Incorrect.
Among these, we identified five sets where all three experts labeled the answers as Perfect. These sets were then used as context examples for the generation model in the main user study.

To validate annotation reliability, we calculated Fleiss’ Kappa across the three experts on a shared annotation subset (10 items) in the preliminary study. Using the original three labels (Accurate, Less accurate, Inaccurate), the agreement reached $\kappa$ = 0.522, indicating moderate agreement despite the finer-grained distinctions required. When consolidating Less accurate and Inaccurate into a single category (Not accurate), Fleiss’ Kappa increased substantially to $\kappa$ = 0.864, demonstrating almost perfect agreement among experts regarding factual correctness. These results confirm that the experts exhibit strong consistency in our ground-truth. 


\subsection{Model selection}\label{model-selection}
To determine the best model for identifying protein-protein interactions, we evaluated four large language models: GPT-4o \cite{achiam2023gpt}, gemini-2.0-flash \cite{google2025gemini2flash}, grok-2-1212 \cite{grok2024}, and deepseek-v3 \cite{deepseek2024}. Each model was tasked with answering three sequential questions that were used in the preliminary user study based on ten abstracts. 


To assess model performance, we compared their predictions to the ten abstracts using several metrics: \textbf{BERTScore}~\cite{zhang2019bertscore}, \textbf{ROUGE-1}, and \textbf{ROUGE-L}~\cite{lin2004rouge}. Overall, GPT-4o was judged the most effective model due to its consistently strong performance and clear, concise responses. As summarized in Table \ref{tab:results}, gpt-4o led in BERT F1, grok-2-1212 topped ROUGE-L F1 and BERT Precision, and deepseek-v3 ranked highest in ROUGE-1 F1 and BERT Recall.

\subsection{Main user study}\label{main_study}
We recruited 10 professional drug discovery researchers. All participants were researching Target ID. These participants were newly recruited for the \system evaluation, distinct from the formative study. Their career experience ranged from 5 to 18 years (mean=13.5, SD=4.3). Our study was approved by the Institutional Review Board (IRB). Each participant received a \$100 gift certificate.
Ten experts were asked to label and revise (if necessary) 50 sets of answers as Perfect, Acceptable, or Incorrect. Figure~\ref{fig:google_form} shows the examples of the questionnaires used in this user study.

\begin{figure*}
    \centering
    \includegraphics[width=1\linewidth]{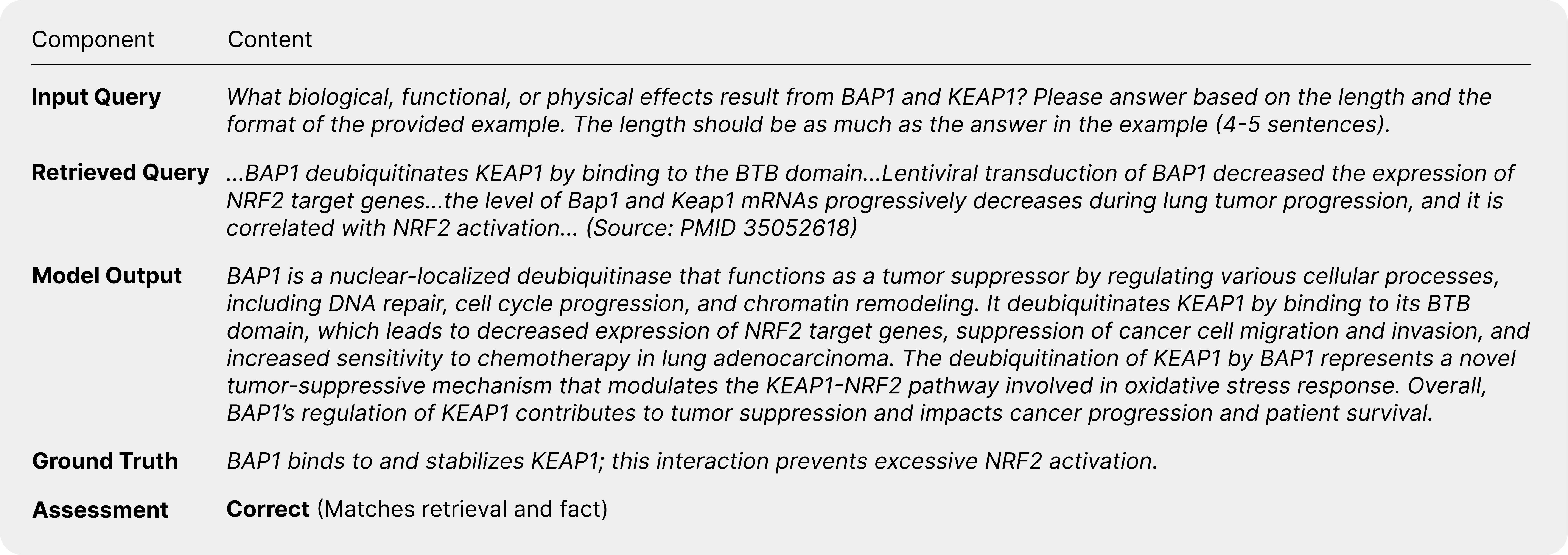}
    \caption{An step-by-step example of input queries, model output, and assessment results.}
    \label{fig:query_example}
\end{figure*}

\clearpage
\captionsetup{type=table,skip=0pt}%
\captionof{table}{The prompt template given to systems to generate results for the preliminary study, the main user study, and silver dataset generation. Examples are edited and validated by domain experts in drug discovery. LLM-based systems do not have access to retrieved context.
}\captionsetup[table]{skip=0pt}
\label{tab:Prompt}
\vspace{-30em}
\begin{mytextbox}
\RaggedRight

\textbf{Question:} \par
"According to the abstract, what biological, functional, or physical effects result from \textit{\{\$ppi\}}? Please answer based on the length and format of the provided example."

\vspace{1em}
\textbf{Abstract:} \par
\textit{\{\$abstract\}}

\vspace{1em}
\textbf{Examples:}
\vspace{0.5em}

\hrule 
\vspace{0.5em}

\textbf{Example-1} \par
\vspace{0.5em}
\textbf{Abstract:} \par
Missense variants located in the N-terminal region of WDR37 were recently identified to cause a multisystemic syndrome affecting neurological, ocular, gastrointestinal, genitourinary, and cardiac development. WDR37 encodes a WD40 repeat-containing protein of unknown function. We identified three novel WDR37 variants, two likely pathogenic de novo alleles and one inherited variant of uncertain significance, in individuals with phenotypes overlapping those previously reported but clustering in a different region of the protein. The novel alleles are C-terminal to the prior variants and located either within the second WD40 motif (c.659A>G p.(Asp220Gly)) or in a disordered protein region connecting the second and third WD40 motifs (c.778G>A p.(Asp260Asn) and c.770C>A p.(Pro257His)). The three novel mutants showed normal cellular localization but lower expression levels in comparison to wild-type WDR37. To investigate the normal interactions of WDR37, we performed co-immunoprecipitation and yeast two-hybrid assays. This revealed the ability of WDR37 to form homodimers and to strongly bind PACS1 and PACS2 phosphofurin acidic cluster sorting proteins; immunocytochemistry confirmed colocalization of WDR37 with PACS1 and PACS2 in human cells. Next, we analyzed previously reported and novel mutants for their ability to dimerize with wild-type WDR37 and bind PACS proteins. Interaction with wild-type WDR37 was not affected for any variant; however, one novel mutant, p.(Asp220Gly), lost its ability to bind PACS1 and PACS2. In summary, this study presents a novel region of WDR37 involved in human disease, identifies PACS1 and PACS2 as major binding partners of WDR37 and provides insight into the functional effects of various WDR37 variants.

\vspace{0.5em}
\textbf{Question 1:} According to the abstract, what biological, functional, or physical effects result from the PPI WDR37>PACS1? \par
\vspace{0.5em}
\textbf{Answer-1:} WDR37, a WD40 repeat-containing protein, forms homodimers and binds strongly to PACS1 and PACS2 phosphofurin acidic cluster sorting proteins. However, a novel mutant variant of WDR37, p.(Asp220Gly), loses its ability to bind to PACS1 and PACS2. This loss of interaction could potentially disrupt normal cellular processes, leading to the multisystemic syndrome affecting various aspects of human development, as observed in individuals with this mutation.

\vspace{1em}
\hrule
\vspace{0.5em}

\textbf{Example-2} \par
\vspace{0.5em}
\textbf{Abstract:} \par
Menin is a 70-kDa protein encoded by MEN1, the tumor suppressor gene disrupted in multiple endocrine neoplasia type 1. In a yeast two-hybrid system based on reconstitution of Ras signaling, menin was found to interact with the 32-kDa subunit (RPA2) of replication protein A (RPA), a heterotrimeric protein required for DNA replication, recombination, and repair. The menin-RPA2 interaction was confirmed in a conventional yeast two-hybrid system and by direct interaction between purified proteins. Menin-RPA2 binding was inhibited by a number of menin missense mutations found in individuals with multiple endocrine neoplasia type 1, and the interacting regions were mapped to the N-terminal portion of menin and amino acids 43 to 171 of RPA2. This region of RPA2 contains a weak single-stranded DNA-binding domain, but menin had no detectable effect on RPA-DNA binding in vitro. Menin bound preferentially in vitro to free RPA2 rather than the RPA heterotrimer or a subcomplex consisting of RPA2 bound to the 14-kDa subunit (RPA3). However, the 70-kDa subunit (RPA1) was coprecipitated from HeLa cell extracts along with RPA2 by menin-specific antibodies, suggesting that menin binds to the RPA heterotrimer or a novel RPA1-RPA2-containing complex in vivo. This finding was consistent with the extensive overlap in the nuclear localization patterns of endogenous menin, RPA2, and RPA1 observed by immunofluorescence.

\vspace{0.5em}
\textbf{Question 2:} According to the abstract, what biological, functional, or physical effects result from the PPI RPA2>MEN1? \par
\vspace{0.5em}
\textbf{Answer-2:} Menin, a tumor suppressor protein, interacts with the RPA2 subunit of the replication protein A (RPA), a protein essential for DNA replication, recombination, and repair. This interaction is disrupted by certain missense mutations in menin found in individuals with multiple endocrine neoplasia type 1. Despite menin's lack of effect on RPA-DNA binding in vitro, it preferentially binds to free RPA2 and may interact with the RPA heterotrimer or a novel RPA1-RPA2-containing complex in vivo, suggesting a potential role in regulating DNA replication and repair processes.

\vspace{1em}
\hrule
\vspace{0.5em}

\textbf{Example-3} \par
\vspace{0.5em}
\textbf{Abstract:} \par
FHL1 has been recognized for a long time as a tumor suppressor protein that associates with both the actin cytoskeleton and the transcriptional machinery. We present in this study a paradigm that phosphorylated FHL1 functions as an oncogenic protein by promoting tumor cell proliferation. The cytosolic tyrosine kinase Src interacts with and phosphorylates FHL1 at Y149 and Y272, which switches FHL1 from a tumor suppressor to a cell growth accelerator. Phosphorylated FHL1 translocates into the nucleus, where it binds to the transcription factor BCLAF1 and promotes tumor cell growth. Importantly, the phosphorylation of FHL1 is increased in tissues from lung adenocarcinoma patients despite the down-regulation of total FHL1 expression. Kindlin-2 was found to interact with FHL1 and recruit FHL1 to focal adhesions. Kindlin-2 competes with Src for binding to FHL1 and suppresses Src-mediated FHL1 phosphorylation. Collectively, we demonstrate that FHL1 can either suppress or promote tumor cell growth depending on the status of the sites for phosphorylation by Src.

\vspace{0.5em}
\textbf{Question 3:} According to the abstract, what biological, functional, or physical effects result from the PPI FHL1>BCLAF1? \par
\vspace{0.5em}
\textbf{Answer-3:} The interaction between FHL1 and BCLAF1, facilitated by the phosphorylation of FHL1 by Src, promotes tumor cell growth. This phosphorylation causes FHL1 to translocate into the nucleus and bind to the transcription factor BCLAF1, switching its role from a tumor suppressor to a cell growth accelerator. However, Kindlin-2 can suppress this process by competing with Src for binding to FHL1, thereby inhibiting FHL1 phosphorylation and its subsequent oncogenic activity.

\vspace{1em}
\hrule
\vspace{0.5em}

\textbf{Example-4} \par
\vspace{0.5em}
\textbf{Abstract:} \par
The Kar3 protein (Kar3p), a protein related to kinesin heavy chain, and the Cik1 protein (Cik1p) appear to participate in the same cellular processes in S. cerevisiae. Phenotypic analysis of mutants indicates that both CIK1 and KAR3 participate in spindle formation and karyogamy. In addition, the expression of both genes is induced by pheromone treatment. In vegetatively growing cells, both Cik1::beta-gal and Kar3::beta-gal fusions localize to the spindle pole body (SPB), and after pheromone treatment both fusion proteins localize to the spindle pole body and cytoplasmic microtubules. The dependence of Cik1p and Kar3p localization upon one another was investigated by indirect immunofluorescence of fusion proteins in pheromone-treated cells. The Cik1p::beta-gal fusion does not localize to the SPB or microtubules in a kar3 delta strain, and the Kar3p::beta-gal fusion protein does not localize to microtubule-associated structures in a cik1 delta strain. Thus, these proteins appear to be interdependent for localization to the SPB and microtubules. Analysis by both the two-hybrid system and co-immunoprecipitation experiments indicates that Cik1p and kar3p interact, suggesting that they are part of the same protein complex. These data indicate that interaction between a putative kinesin heavy chain-related protein and another protein can determine the localization of motor activity and thereby affect the functional specificity of the motor complex.

\vspace{0.5em}
\textbf{Question 4:} According to the abstract, what biological, functional, or physical effects result from the PPI CIK1>KAR3? \par
\vspace{0.5em}
\textbf{Answer-4:} The interaction between Cik1p and Kar3p, both of which are involved in spindle formation and karyogamy, determines their localization to the spindle pole body (SPB) and microtubules. In the absence of either protein, the other fails to localize to these structures, indicating their interdependence. This interaction likely forms a protein complex that influences the localization of motor activity, thereby affecting the functional specificity of the motor complex.

\vspace{1em}
\hrule
\vspace{0.5em}

\textbf{Example-5} \par
\vspace{0.5em}
\textbf{Abstract:} \par
The MYC oncoprotein, also known as the master regulator of genes, is a transcription factor that regulates numerous physiological processes, including cell cycle control, apoptosis, protein synthesis and cell adhesion, among others. MYC is overexpressed in approximately 70\% of human cancers. Given its pervasive role in cancer biology, MYC down-regulation has become an attractive cancer treatment strategy.Methods:The CRISPR/Cas9 method was used to produce KO cell models. Western blot was used to analyzed the expressions of MYC and TATA-binding proteinassociated factors 10 (TAF10) in cancer cells (MCF7, A549, HepG2 cells) Cell culture studies were performed to determine the mechanisms by which small molecules (Z363119456, Z363) affects MYC and TAF10 expressions and functions. Mouse studies were carried out to investigate the impact of Z363 regulation on tumor growth.Results:Z363 activate Thyroid hormone Receptor-interacting Protein 12 (TRIP12), which phosphorylates MYC at Thr58, resulting in MYC ubiquitination and degradation and thereby regulating MYC target genes. Importantly, TRIP12 also induces TAF10 degradation, which reduces MYC protein levels. TRIP12, an E3 ligase, controls MYC levels both directly and indirectly by inhibiting MYC or TAF10 activity.Conclusions:In summary,these results demonstrate the anti-cancer properties of Z363, a small molecule that is co-regulated by TAF10 and MYC.

\vspace{0.5em}
\textbf{Question 5:} According to the abstract, what biological, functional, or physical effects result from the PPI TRIP12>TAF10? \par 
\vspace{0.5em}
\textbf{Answer-5:} The activation of TRIP12 by Z363 leads to the phosphorylation of MYC at Thr58, which triggers its ubiquitination and subsequent degradation, thereby regulating MYC target genes. Concurrently, TRIP12 also induces the degradation of TAF10, which further reduces MYC protein levels. This dual action of TRIP12, both directly and indirectly, controls MYC levels and inhibits its activity, demonstrating the potential anti-cancer properties of Z363. 

\end{mytextbox}

\end{document}